\pdfoutput=1
\documentclass[11pt]{article}

\usepackage{acl}
\usepackage{times}
\usepackage{latexsym}

\usepackage[T1]{fontenc}
\usepackage[utf8]{inputenc}
 \usepackage{array,multirow,graphicx}
 \usepackage{float}
\usepackage{microtype}

\usepackage{inconsolata}
\usepackage{xspace}
\usepackage{paralist}
\newif{\ifhidecomments}
\hidecommentsfalse
\newcommand{\catwalk}{Catwalk\xspace}
\newcommand{\resolved}[1]{}

\usepackage{enumitem}
\usepackage{graphicx}

\usepackage{amssymb}% http://ctan.org/pkg/amssymb
\usepackage{pifont}% http://ctan.org/pkg/pifont
\newcommand{\cmark}{\ding{51}}%
\newcommand{\xmark}{\ding{55}}%

\usepackage{rotating}
\usepackage{array}
\usepackage{booktabs}  
\newcolumntype{R}[1]{>{\begin{turn}{90}\begin{minipage}{#1}\scriptsize}l%
<{\end{minipage}\end{turn}}%
}
\usepackage{colortbl}
\usepackage{caption}
\usepackage{subcaption}

\usepackage{listings}
\usepackage{xcolor}

\definecolor{eclipseStrings}{RGB}{42,0.0,255}
\definecolor{eclipseKeywords}{RGB}{127,0,85}
\colorlet{numb}{magenta!60!black}

\lstdefinelanguage{json}{
    basicstyle=\footnotesize\ttfamily,
    commentstyle=\color{eclipseStrings}, % style of comment
    stringstyle=\color{eclipseKeywords}, % style of strings
    numbers=left,
    numberstyle=\scriptsize,
    stepnumber=1,
    numbersep=8pt,
    showstringspaces=false,
    breaklines=true,
    frameround=tttt,
    frame = single,
    string=[s]{"}{"},
    comment=[l]{\\},
    morecomment=[l]{:"},
    morekeywords=[2]{local,import},
    morekeywords=[3]{
        answers,
        choices,
        RankClassificationInstance,
        correct_choice,
        label,
        text,
        context,
        id,
        HFMCInstance,
        HFQAInstance,
        HFClassificationInstance,
        question,
        answer_choices,
        correct_answer_index,
        model_path,
        revision,
        gpus_needed,
        prediction_kwargs,
        model_max_length,
        max_batch_tokens,
        steps
    },
    keywordstyle=[2]\color{red},
    keywordstyle=[3]\color{eclipseStrings},
    literate=
        *{0}{{{\color{numb}0}}}{1}
         {1}{{{\color{numb}1}}}{1}
         {2}{{{\color{numb}2}}}{1}
         {3}{{{\color{numb}3}}}{1}
         {4}{{{\color{numb}4}}}{1}
         {5}{{{\color{numb}5}}}{1}
         {6}{{{\color{numb}6}}}{1}
         {7}{{{\color{numb}7}}}{1}
         {8}{{{\color{numb}8}}}{1}
         {9}{{{\color{numb}9}}}{1}
}
\title{Catwalk: A Unified Language Model Evaluation Framework\\for Many Datasets}

\author{
  \textbf{Dirk Groeneveld}$^\spadesuit$ \quad
  \textbf{Anas Awadalla}$^{\diamondsuit}$ \quad
  \textbf{Iz Beltagy}$^\spadesuit$ \quad
  \textbf{Akshita Bhagia}$^\spadesuit$ \quad\\
  \textbf{Ian Magnusson}$^\spadesuit$ \quad
  \textbf{Hao Peng}$^\spadesuit$ \quad 
  \textbf{Oyvind Tafjord}$^\spadesuit$ \quad
  \textbf{Pete Walsh}$^\spadesuit$ \quad\\
  \textbf{Kyle Richardson}$^\spadesuit$ \quad
  \textbf{Jesse Dodge}$^\spadesuit$ \\
  $^\spadesuit$Allen Institute for Artificial Intelligence \\
  $^{\diamondsuit}$Paul G. Allen School of Computer Science \& Engineering,
  University of Washington \\
   {\tt anasa2@cs.washington.edu} \\
   {\tt \{dirkg,beltagy,akshitab,ianm,haop,oyvindt,petew,kyler,jessed\}@allenai.org}
}
  
\begin{document}
% \raggedbottom
\maketitle
\begin{abstract}

The success of large language models  has shifted the evaluation paradigms in natural language processing (NLP).
The community's interest has drifted towards comparing NLP models across \emph{many} tasks, domains, and datasets, often at an extreme scale.
This imposes new engineering challenges: efforts in constructing datasets and models have been fragmented, and their formats and interfaces are incompatible.
As a result, it often takes extensive (re)implementation efforts to make fair and controlled comparisons at scale.

\catwalk aims to address these issues.
\catwalk provides a unified interface to a broad range of existing NLP datasets and models, ranging from both canonical supervised training and fine-tuning, to more modern paradigms like in-context learning.
Its carefully-designed abstractions allow for easy extensions to many others.
% \catwalk is efficient. \hao{i want to talk about the caching thing by Tango in one sentence, but I'm not sure i know enough about tango to achieve this. anyone can help?}
\catwalk substantially lowers the barriers to conducting controlled experiments at scale.
For example, we finetuned and evaluated over 64 models on over 86 datasets with a single command, \emph{without writing any code}.
Maintained by the AllenNLP team at the Allen Institute for Artificial Intelligence (AI2),
\catwalk is an ongoing open-source effort: \url{https://github.com/allenai/catwalk}.
\end{abstract}

\section{Introduction}

\begin{figure}[!t]
	\centering
\begin{subfigure}[b]{.48\textwidth}
\centering
	\includegraphics[width=\columnwidth, trim={14cm 11cm 14cm 12cm},clip]{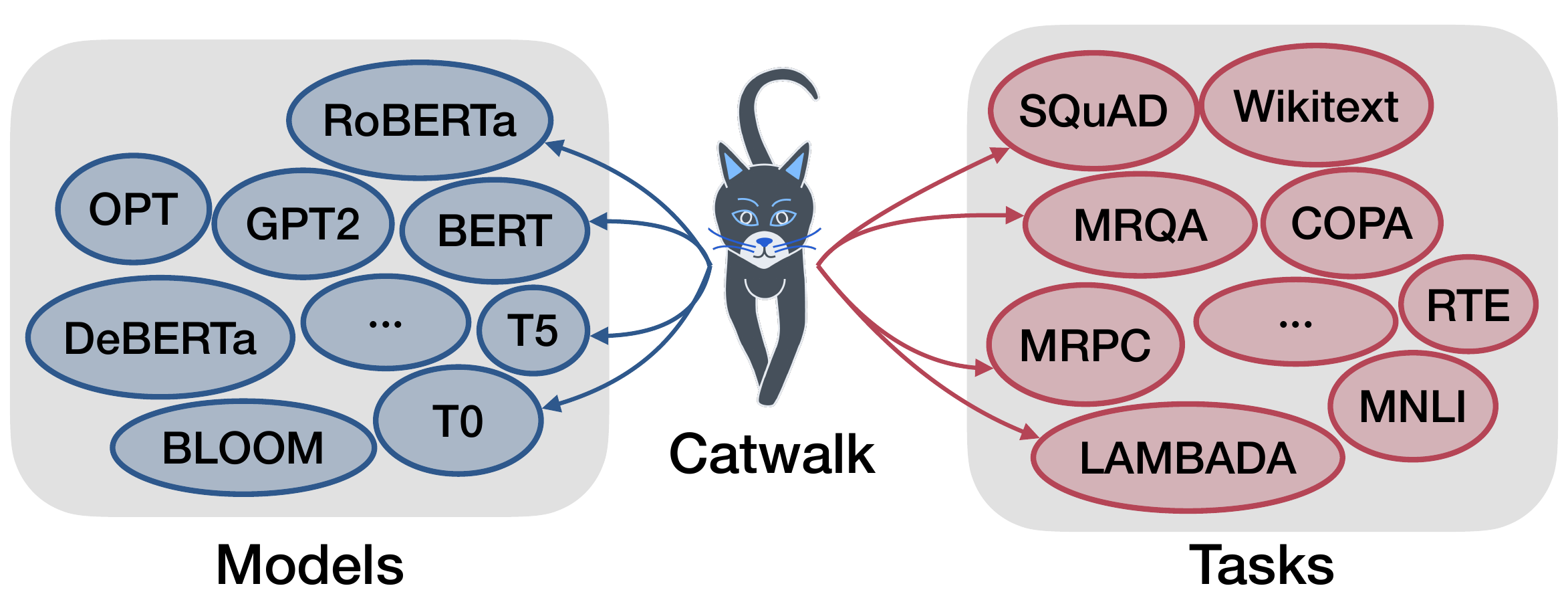}
    \caption{\catwalk provides a unified interfaces to connect models and datasets. This compares $n$ models on $m$ tasks with only $n + m$ custom implementations, substantially reducing the implementation workload compared to current practice where $nm$ implementations are needed.}
	\label{fig:diagram}
 \end{subfigure}
 \hfill
 \begin{subfigure}[b]{.48\textwidth}
 \centering
	\includegraphics[width=\columnwidth, page=1]{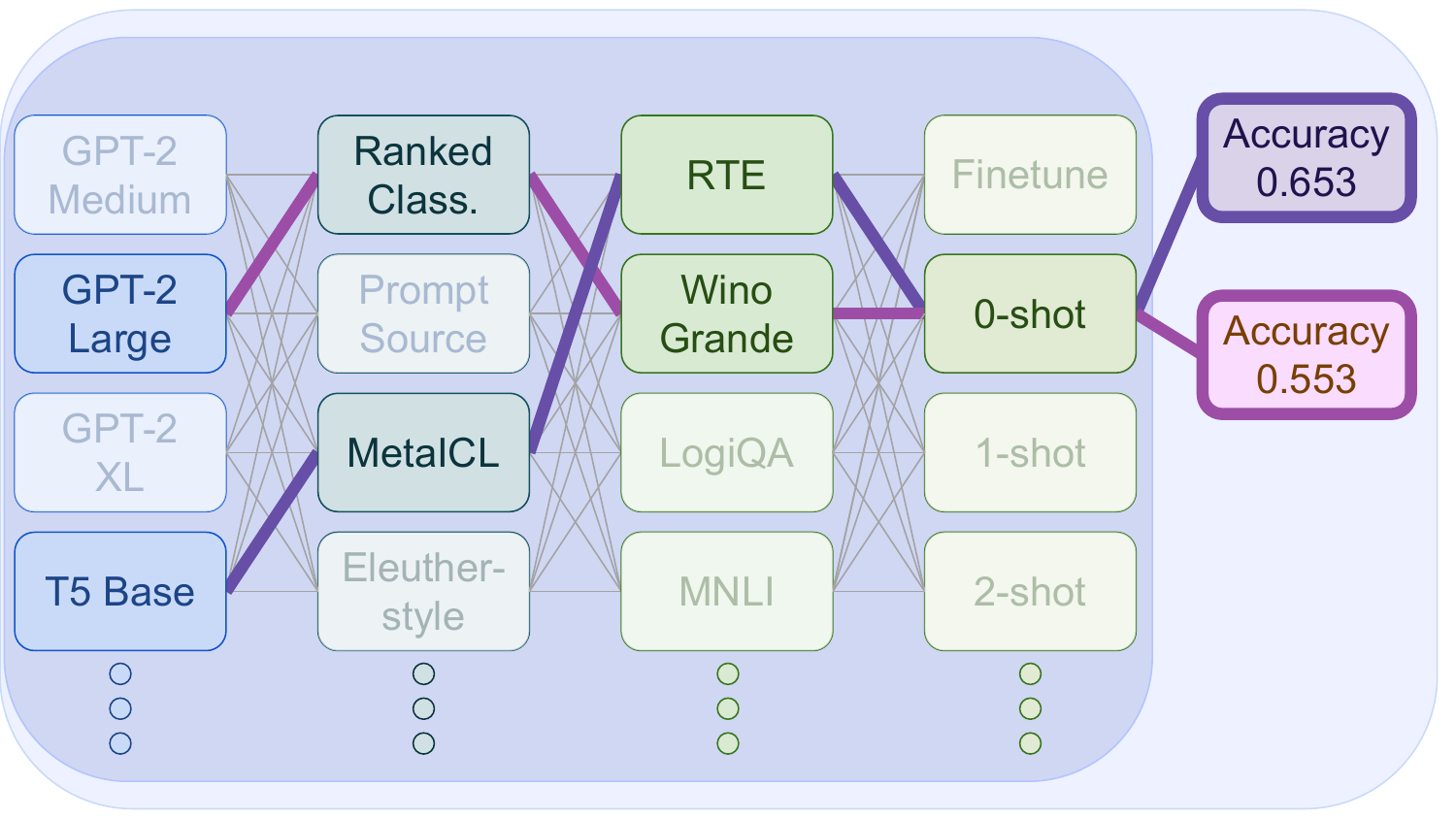}
        	\caption{Illustration of some decisions a user of \catwalk makes to get a result. For example, choosing GPT-2 Large, with Ranked Classification, on WinoGrande gives an accuracy of $0.553$, while choosing T5-Base on RTE gives an accuracy of $0.653$.
        }
	\label{fig:overview}
 \end{subfigure}
 \caption{\catwalk and its workflow.}
\end{figure}

Large language models (LLMs) have led to a paradigm shift in NLP.
Before, a system would be built to tackle a specific task and trained and evaluated on a dedicated dataset.
Now, general-purpose pretrained language models are adapted to a broad range of tasks through techniques like finetuning and in-context learning.
The new paradigm opens research opportunities to investigate models' generalization and robustness across \emph{many} tasks and domains. 
This imposes new engineering challenges.
The fragmented efforts in producing models and datasets have led to countless codebases and data formats that are hardly compatible.
Fairly comparing $n$ models on $m$ datasets often require $nm$ custom implementations, substantially raising the barriers to conducting controlled experiments at scale.

\catwalk aims to reduce practitioners' workload in projects involving making fair comparisons among many models and datasets.
At a high level, \catwalk can be thought of as a nexus connecting $n$ models and $m$ datasets.
Through Catwalk's unified interfaces, training and evaluating all models on all datasets requires $n + m$ customizations instead of $mn$, many of which have already been implemented.
To include a new model or dataset into \catwalk, one only needs to instantiate a few interfaces, and it can be evaluated on \emph{all} datasets/models that \catwalk supports. Figures~\ref{fig:diagram} and \ref{fig:overview} provide illustrative examples.
Whenever possible, \catwalk caches and reuses intermediate experimental steps such as loaded datasets/models, model predictions, as well as trained models.\footnote{Catwalk's caching mechanism is enabled by AI2-Tango~\citep{tango}: \url{https://github.com/allenai/tango}.}
This helps minimize redundant machine workloads in, e.g., loading the data and training the models, and better utilize the limited hardware resources.
\catwalk's caching mechanism is especially useful in large-scale experiments that involve many models and/or many tasks.

\catwalk is an ongoing effort. We aim to jump-start a virtuous cycle of dataset and model authors adopting support for \catwalk's unified interfaces. 
To this end, we contribute:
\begin{compactitem}
    \item 6 model abstractions that support a variety of existing models.
    \item Zero-shot and few-shot evaluations, as well as fine-tuning.
    \item 9 data formats that cover common use cases.
    \item Support for 86 standalone datasets as well as curated benchmarks that contain many datasets themselves, for a total of over 800 datasets, along with evaluation metrics.
    \item A case study that compares 30 widely-used models on 17 datasets.
\end{compactitem}
We expect the above list to keep growing.
% \catwalk is maintained by the AllenNLP team at Allen Institute for Artificial Intelligence (AI2).
\catwalk is open-source: \url{https://github.com/allenai/catwalk}.
It is maintained by the AllenNLP team at the Allen Institute for Artificial Intelligence (AI2)
and has been a useful tool for their projects, including being the main evaluation framework used in the OLMo (Open Language Model) project.\footnote{See more details at \url{https://allenai.org/olmo} as well as \url{https://github.com/allenai/ai2-olmo-eval} for details about our LLM evaluation efforts.}
We hope it can benefit the community's research too. \resolved{\hao{i don't have strong feelings about this wording. feel free to drop if anyone is uncomfortable with saying this}}

\section{Structure and organization}
The purpose of \catwalk is to answer the question "How well does model X do on dataset Y?", for as many models and datasets as possible. The possible models include GPT-style decoder-only models~\cite{gpt}, T5-style encoder/decoder models~\cite{t5}, and BERT-style fine-tuned models~\cite{bert}. \catwalk imposes no constraint on the types of models it can support. 
The model does not have to be transformer-based or even be a language model at all, though all currently implemented models are transformer-based language models.

To achieve this flexibility, \catwalk 
%offers the ability to 
transforms instances from a dataset into ``formats'' that are more suited to model execution. Crucially, these formats are the same across many datasets, so model implementations do not have to concern themselves with the differences between datasets, and can be written in a simple, generic style.

\begin{table*}[t]
\centering
\begin{tabular}{@{} l c c c c c c @{}}
    \toprule
    Model style & Enc-only & Dec-only & Enc/dec & Finetuning & Zero-shot & Few-shot \\
    \midrule
    Ranked classification & \xmark & \cmark & \cmark & \cmark & \cmark & \cmark \\
    Bert-style & \cmark & \xmark & \xmark & \cmark & \xmark & \xmark \\
    Language models & \xmark & \cmark & \xmark & \xmark & \cmark & \cmark \\ \hline
    PromptSource & \xmark & \cmark & \cmark & \cmark & \cmark & \cmark \\
    MetaICL & \xmark & \cmark & \cmark & \xmark & \cmark & \cmark \\
    Eleuther-style & \xmark & \cmark & \xmark & \xmark & \cmark & \cmark \\
    \bottomrule
\end{tabular}
\caption{Model styles and supported features in \catwalk.}
\vspace{-.5cm}
\label{tab:wrappers}
\end{table*}

\section{Models and Model Wrappers}

\catwalk's models combine an underlying model, a model wrapper, and a method to process the outputs from the model. For example, a decoder-only language model can be used to compute the probability of various answer options. This can be done using T0-style human-readable prompts~\cite{t0}, or it can be done using machine-readable prompts in a few-shot setting~\cite{eval-harness}. This model could even be fine-tuned and used with a task-specific head~\cite{bert}. These options have different characteristics and produce different results. They are all encapsulated in \catwalk's model wrapper abstraction.

\subsection{Model Wrappers}

Model wrappers come in two general varieties, \emph{native interfaces} that provide generic support for a wide class of models and \emph{adapter interfaces} that aim to facilitate systematic comparisons with existing model types or evaluation frameworks. We list the different model wrappers below current implemented in Catwalk (with the first three being our native interfaces and the last three being example adapter interfaces; see Table~\ref{tab:wrappers}).

\begin{compactitem}
    \item \textbf{Ranked classification models}. This model wrapper uses decoder-only and encoder/decoder models. It calculates the probability of all answer options, and chooses the most likely answer. It supports zero-shot and few-shot evaluations, and it can be fine-tuned on a dataset, or a list of datasets.
    \item \textbf{BERT-style models}. This model wrapper uses the native API from the Huggingface library~\cite{huggingface} to perform multiple-choice, question answering, and classification tasks. This is the way to run encoder-only models like BERT~\cite{bert}, RoBERTa~\cite{roberta}, or DeBERTa~\cite{deberta}. It supports fine-tuning, but no few-shot evaluations.
    \item \textcolor{black}{\textbf{Language models}. This model wrapper is a generic class for decoder-only generation models that supports few- and zero-shot evaluation. This is the class most relevant to recent GPT-style LLM evaluation. In contrast to Eleuther-style models \cite{eval-harness} \emph{(see below)}, it is designed to support a wider range of generation metrics and token-based scoring strategies beyond those hard-coded in existing toolkits. It also has special features for returning per-instance predictions that allow for more fine-grained analysis and debugging.}
    \item \textbf{PromptSource models}. This model wrapper also performs ranked classification, but it uses the human-readable prompts from the PromptSource project~\cite{promptsource}. It supports zero-shot and few-shot evaluations.
    \item \textbf{MetaICL models}. This model wrapper applies the prompt truncation strategy used by \citet{min-etal-2022-metaicl} to decoder-only ranked classification. It truncates each of several in-context demonstrations separately and truncates again after concatenation.
    \item \textbf{Eleuther-style models}. This model wrapper uses decoder-only models in the style of \citet{eval-harness}. It reproduces those numbers exactly, though it runs a little faster. Like \citet{eval-harness}, it supports zero-shot and few-shot evaluations. On many datasets, this is equivalent to the ranked classification models.
\end{compactitem}

\subsection{Models}
All of \catwalk's existing model implementations load parameters and configurations from Huggingface. They can be configured to load any model from the Huggingface hub, but \catwalk gives a few of them names, tests them regularly, and makes them available on the command line. The following is a list of models that have names in \catwalk and that have been extensively tested:

\begin{itemize}[noitemsep,topsep=0pt,parsep=0pt,partopsep=0pt]
\item BERT \{tiny, base, large\}, \{cased, uncased\}~\cite{bert}.
\item BLOOM \{560M, 1B1, 1B7, 3B, 7B1\}~\cite{bloom}.
\item CT0-11b~\cite{ct0}.
\item DEBERTA \{v3-small, v3-base, v3-large, v2-xlarge, v2-xxlarge\}~\cite{deberta,he2021debertav3}.
\item DistilBERT-base-cased-distilled-squad~\cite{distilbert}.
\item GPT-J-6B~\cite{gptj}.
\item GPT-Neo \{125M, 1.3B, 2.7B\}~\cite{pile}.
\item GPT-NeoX-20B~\cite{gpt-neox-20b}.
\item GPT2 \{tiny, regular, medium, large, xl\}~\cite{gpt}.
\item mT5 \{small, base, large, xl\}~\cite{mt5}.
\item OPT \{125M, 350M, 1.3B, 2.7B, 6.7B, 13B, 30B, 66B\}~\cite{opt}.
\item RoBERTa \{base, large\}~\cite{roberta}.
\item T0 \{regular, 3B, original\_task\_only, single\_prompt, p, pp\}~\cite{t0}.
\item T5 \{very\_small\_random, small, small-lm-adapt, v1\_1-small, base, base-lm-adapt, v1\_1-base, large, large-lm-adapt, v1\_1-large, 3B, xl-lm-adapt, v1\_1-xl, 11b, xxl-lm-adapt, v1\_1-xxl\}~\cite{t5}.
\item Pythia \{70M,160M,410M,1B,1.4B,2.7B,6.9B,12B\} \cite{biderman2023pythia}
\item Falcon \{1B,7B,40B\}~\cite{falcon40b}
\item Llama(-2) \{7B, 13B, 70B\}~\cite{touvron2023llama,touvron2023llama2}
\item MPT \{7B, 30B\}~\cite{MosaicML2023Introducing}
\end{itemize}

\section{Prompting}

Catwalk supports a number of formats for dataset instances.
By converting dataset instances into a common format, we allow model implementations to only depend on the format, and not on the vagaries of individual datasets.

Catwalk supports the following formats.

\subsection{Huggingface dictionaries}

This format makes few guarantees beyond being a Python dictionary.
It is the raw format of the data, and rarely useful directly.
This is the format that the Huggingface datasets library~\cite{datasets} produces.
It isn't used by any of the models directly, but all the other formats use this as their starting point.
\noindent\begin{minipage}{\linewidth}
\begin{lstlisting}[language=json,frame=single,label=hf_dict,basicstyle=\tiny,showstringspaces=false,numbers=none]
{
  "sentence1": "No Weapons of Mass Destruction Found Yet.",
  "sentence2": "Weapons of Mass Destruction Found.",
  "label": 1,
  "idx": 0
}
\end{lstlisting}
\end{minipage}

\subsection{Huggingface multiple choice}

This is a structured format containing a question, a list of answer choices, and optionally the index of the correct answer.
This corresponds exactly to the expectation of Huggingface multiple choice models.
It is used by the BERT-style models.
\noindent\begin{minipage}{\linewidth}
\begin{lstlisting}[language=json,frame=single,label=hf_mc,basicstyle=\tiny,showstringspaces=false,numbers=none]
HFMCInstance(
  id="Mercury_7220990",
  question="Which factor will most likely cause a " +
    "person to develop a fever?",
  answer_choices=[
    "a leg muscle relaxing after exercise",
    "a bacterial population in the bloodstream",
    "several viral particles on the skin",
    "carbohydrates being digested in the stomach"
  ],
  correct_answer_index=1
)
\end{lstlisting}
\end{minipage}

\subsection{Huggingface question-answering}

This is a structured format containing a context, a question, and a list of possible correct answers.
This corresponds exactly to the expectation of Huggingface question-answering models.
It is used by the BERT-style models.

\noindent\begin{minipage}{\linewidth}
\begin{lstlisting}[language=json,frame=single,label=hf_mc,basicstyle=\tiny,showstringspaces=false,numbers=none]
HFQAInstance(
  id="5733be284776f41900661182",
  question="To whom did the Virgin Mary appear in 1858?",
  context="... the Virgin Mary appeared to Saint " +
    "Bernadette Soubirous in 1858 ...",
  answers={
    "text": ["Saint Bernadette Soubirous"],
    "answer_start": [515]})
\end{lstlisting}
\end{minipage}

\subsection{Huggingface classification}

This is a structured format containing a context, as a string or a pair of strings, and an optional label.
The possible answer choices are not part of the format but are part of the model that uses these instances.
This corresponds exactly to the expectation of Huggingface classification models.
It is used by the BERT-style models.
\noindent\begin{minipage}{\linewidth}
\centering
\begin{lstlisting}[language=json,frame=single,label=hf_mc,basicstyle=\tiny,showstringspaces=false,numbers=none]
HFClassificationInstance(
  text=(
    "No Weapons of Mass Destruction Found Yet.",
    "Weapons of Mass Destruction Found."
  ), label=1)
\end{lstlisting}
\end{minipage}
\subsection{Eleuther}

There are several formats specific to the Eleuther evaluation framework~\cite{eval-harness}.
They are useful to build other format converters on top of, but are rarely useful on their own.
Only the Eleuther-style models use these formats directly.

\subsection{T5}

This format represents instances as strings starting with the dataset name followed by named fields that make up the instance.
It also includes a gold answer that is expected of the model.
This corresponds exactly to the way T5~\cite{t5} was trained.
The format is only supported for a small number of datasets.
\noindent\begin{minipage}{\linewidth}
\begin{lstlisting}[language=json,frame=single,label=hf_mc,basicstyle=\tiny,showstringspaces=false,numbers=none]
(
 "rte sentence1: No Weapons of Mass Destruction found yet
    sentence2: Weapons of Mass Destruction Found.",
   "not_entailment"
)
 \end{lstlisting}
 \end{minipage}

\subsection{Ranked classification}

This format represents instances as a list of options, together with an optional label indicating the correct option.
Each option is a text pair, called ``context'' and ``continuation''.
Models generally compute the probability of the continuation given the context to determine the best answer, though recent work suggests other possibilities~\cite{channel}.
The format is used by the built-in ranked classification models, and it forms the basis for the PromptSource format.
\noindent\begin{minipage}{\linewidth}
\begin{lstlisting}[language=json,frame=single,label=hf_mc,basicstyle=\tiny,showstringspaces=false,numbers=none]
RankClassificationInstance(
  choices=[
    (
      "No Weapons of Mass Destruction Found Yet.\n" +
        "Question: Weapons of Mass Destruction Found. " +
        "True or False?\nAnswer:",
      " True"
    ), (
      "No Weapons of Mass Destruction Found Yet.\n" +
        "Question: Weapons of Mass Destruction Found. " +
        "True or False?\nAnswer:",
      " False")
  ],correct_choice=1)
\end{lstlisting}
\end{minipage}

\subsection{PromptSource}

This format makes the prompts from the PromptSource project~\cite{promptsource} available in \catwalk.
In the PromptSource project, every dataset has multiple prompts.
These are reflected in the format by having multiple versions of the instance, one per prompt, in a dictionary.
Every possible version looks exactly like a ranked classification instance, and can be used by model implementations accordingly.
The PromptSource model implementation uses this format.

\noindent\begin{minipage}{\linewidth}
\begin{lstlisting}[language=json,frame=single,label=promptsource,basicstyle=\tiny,showstringspaces=false,numbers=none]
{"does the claim": RankClassificationInstance(
    choices=[
      (
        "Does the claim 'Weapons of Mass Destruction " +
          "Found.' follow from the fact that 'No Weapons " +
          "of Mass Destruction Found Yet.'? Please " +
          "answer either yes or no.",
        "yes"
      ), (
        "Does the claim 'Weapons of Mass Destruction '" +
          "Found.' follow from the fact that 'No Weapons " +
          "of Mass Destruction Found Yet.'? Please " +
          "answer either yes or no.", "no"
      )
    ],
    correct_choice=1),
  "imply": RankClassificationInstance(
    choices=[
      (
        "Does 'No Weapons of Mass Destruction Found " +
          "Yet.' imply that 'Weapons of Mass Destruction " +
          "Found.'? Please answer either yes or no.", "yes"
      ), (
        "Does 'No Weapons of Mass Destruction Found " +
          "Yet.' imply that 'Weapons of Mass Destruction " +
          "Found.'? Please answer either yes or no.",
        "no"
      )
    ],correct_choice=1)}
\end{lstlisting}
\end{minipage}

\subsection{Perplexity}

Catwalk also has a general-purpose task format for perplexity evaluation, which was developed as part of the PALOMA benchmark \cite{paloma}. It follows best practices for perplexity analysis such as avoiding document concatenation \cite{pile}, has advanced batching strategies (e.g., sorting by input length to avoid excessive padding) and supports both  non-overlapping and sliding window inference from \citet{press2020shortformer} to handle common issues related to document maximum length truncation.

\section{Datasets}
The datasets supported by \catwalk can be extended with just a few lines of configs with prompts.
Currently \catwalk contains the following datasets:

\paragraph{Perplexity}
\begin{itemize}[noitemsep,topsep=0pt,parsep=0pt,partopsep=0pt]
\item WikiText~\cite{wikitext}
\item LAMBADA~\cite{lambada}
%\item \textcolor{red}{PALOMA?} 
\end{itemize}

\paragraph{Classification}
\begin{itemize}[noitemsep,topsep=0pt,parsep=0pt,partopsep=0pt]
\item CoLA~\cite{warstadt-etal-2019-neural}
\item MRPC~\cite{dolan-brockett-2005-automatically}
\item QQP~\cite{quora-question-pairs}
\item SST-2~\cite{socher-etal-2013-recursive}
\item BoolQ~\cite{clark-etal-2019-boolq}
\item MultiRC~\cite{khashabi-etal-2018-looking}
\item WiC~\cite{pilehvar-camacho-collados-2019-wic}
\item MC-TACO~\cite{mctaco}
\item PubMedQA~\cite{pubmedqa}
\item Hendrycks Ethics~\cite{hendrycks}
\item RAFT~\cite{raft}
\end{itemize}

\paragraph{Open-ended QA}
\begin{itemize}[noitemsep,topsep=0pt,parsep=0pt,partopsep=0pt]
\item SQuAD~\cite{squad}
\item SquadShifts~\cite{squadshifts}
\item MRQA~\cite{mrqa}
\item SQuAD2~\cite{squad2}
\item DROP~\cite{drop}
\item TriviaQA~\cite{triviaqa}
\item WebQuestions~\cite{webqs}
\item TruthfulQA~\cite{truthfulqa}
\item Hendrycks Math~\cite{math}
\item Arithmetic~\cite{BrownGPT32020}
\item Anagrams 1 and 2~\cite{BrownGPT32020}
\item Cycle Letters~\cite{BrownGPT32020}
\item Random Insertion~\cite{BrownGPT32020}
\item Reversed Words~\cite{BrownGPT32020}
\item Natural Questions~\cite{kwiatkowski2019natural}
\end{itemize}

\paragraph{Multiple-choice QA}
\begin{itemize}[noitemsep,topsep=0pt,parsep=0pt,partopsep=0pt]
\item PIQA~\cite{piqa}
\item COPA~\cite{gordon-etal-2012-semeval}
\item WSC~\cite{wsc}
\item PROST~\cite{prost}
\item SciQ~\cite{sciq}
\item QA4MRE~\cite{qa4mre}
\item ARC Easy and Challenge~\cite{arc}
\item LogiQA~\cite{logiqa}
\item HellaSwag~\cite{hellaswag}
\item OpenBookQA~\cite{openbookqa}
\item RACE~\cite{race}
\item HEAD-QA \cite{headqa}
\item MathQA~\cite{mathqa}
\item WINOGRANDE~\cite{winogrande}
\item MuTual~\cite{mutual}
\item MMLU~\cite{hendrycks2020measuring}
\item CaseHOLD~\cite{zheng2021does}
\item SocialIQA~\cite{sap2019socialiqa}
\item CSQA~\cite{talmor2018commonsenseqa}
\end{itemize}

\paragraph{Summarization}
\begin{itemize}[noitemsep,topsep=0pt,parsep=0pt,partopsep=0pt]
\item SciTLDR~\cite{cachola2020tldr}
\item XSUM~\cite{D18-1206}
\item EURLEX~\cite{aumiller2022eur}
\end{itemize}

\paragraph{NLI}
\begin{itemize}[noitemsep,topsep=0pt,parsep=0pt,partopsep=0pt]
\item RTE from GLUE~\cite{glue}
\item RTE from SuperGLUE~\cite{superglue}
\item MNLI~\cite{williams-etal-2018-broad}
\item QNLI~\cite{glue}
\item WNLI~\cite{glue}
\item CB~\cite{Marneffe2019TheCI}
\item Adverserial NLI~\cite{anli}
\end{itemize}

\paragraph{Mixtures of tasks}
\begin{itemize}[noitemsep,topsep=0pt,parsep=0pt,partopsep=0pt]
\item P3~\cite{sanh2022multitask}
\item MetaICL~\cite{min-etal-2022-metaicl}
\end{itemize}

Note that some of these datasets are collections that show up in \catwalk as multiple subsets. For example, there are actually 11 RAFT datasets, each taking a different angle on language model evaluation. In total, this adds up to over 800 datasets.

Adding new datasets to \catwalk is easy. There are classes for including datasets from the Huggingface hub, and format conversion functions can often be written in a single line of code.

\section{Evaluation}

In \catwalk, the responsibility for calculating metrics lies with the model implementations.
Model implementations can calculate any metric that makes sense in the context of the model.
However, for a given dataset we often want the same metrics independent of the model, so we can compare models against each other.
To facilitate this, \catwalk allows one to attach metrics to datasets, which act as a suggestion to the models for which metrics to compute.
All implemented model styles except the Eleuther style compute the suggested metrics.

By default, \catwalk computes the following metrics
and can be easily extended to include others.
\begin{compactitem}
    \item For multiple-choice, classification, and entailment datasets: Accuracy and relative improvement over the random baseline
    \item For question-answering datasets: SQuAD metric as defined by~\citet{squad}.
    \item For language modeling datasets: Perplexity per word, perplexity per byte, and entropy.
\end{compactitem}

\section{Case Studies}
We include some examples of the kinds of analyses \catwalk enables and in the end describe some ongoing work.
Table~\ref{tab:big_matrix} in the Appendix gives an example of the dense matrix of models and datasets that \catwalk can generate. There are many research questions that can be posed and answered with a matrix of results like this:

\paragraph{What strategy for adapting pretrained models to downstream tasks performs best overall?} For instance we can compare finetuning an encoder-only model against zero-shot evaluation on autoregressive models. We find that, macro averaged over 17 datasets, the best finetuned models had 62\% greater relative improvement in accuracy over a random baseline than the zero-shot models , although the largest zero-shot model is an order of magnitude larger than the largest finetuned model. 

\paragraph{Which datasets are easy/hard?}
\emph{All} zero-shot and finetuned models perform well on SciQ, with an average of more than 200\% relative accuracy improvement compared to the random-guess baseline. 
Among the datasets in our experiments, QNLI is the most challenging one for zero-shot ranked classification models, with all models barely outperforming random guesses.
Unsurprisingly, all finetuned models substantially outperform the chance baseline on all datasets, and in all cases they outperform zero-shot models by a wide margin.

\paragraph{What impact does model type or size have across datasets?} Among zero-shot models, we see that, while T5 variants are the only ones that significantly outperform the chance baseline on MNLI, QQP, and RTE,
they \emph{underperform} others and even the random baseline on CoLA and COPA. This could be attributed to the different choices T5 models make in terms of learning objectives, pretraining data, etc., which requires further investigation. Also, in both the zero-shot ranked classification and finetuning settings, larger model sizes, in general, lead to better performance \emph{within the same model family}, with few exceptions.

\paragraph{Is the ordering of models by performance consistent across datasets?} % Each dataset produces a different order. 
Table \ref{tab:rank_correl} in the Appendix shows that every dataset puts the models into roughly the same order by accuracy, but correlations are often weak. The largest correlations are observed between the ARC Easy, ARC Challenge, and OpenbookQA datasets. All three of these are multiple-choice datasets targeting science questions. On the other end of the spectrum, LogiQA and SciQ correlate the least with other datasets, with coefficients below $0.5$. This might be explained by LogiQA being the most difficult, and SciQ being the easiest dataset among the 17 that were part of this study. In one case the models cannot get much signal out of the dataset, and in the other they saturate. A thorough investigation in this style would have to include many more models and prompting methods.

These analyses only scratch the surface of the questions that \catwalk can help answer.
There are many open questions that bear investigation:
\begin{compactitem}
    \item How well does perplexity predict performance on other datasets?
    \item Do the zero-shot results hold in few-shot settings? How does the number of few-shot examples affect the results?
    \item How important is the format of the prompt?
    \item Some large language models make intermediate checkpoints available for analysis. At what point in model training do abilities emerge?
\end{compactitem}

\paragraph{Ongoing work} \textcolor{black}{As mentioned at the outset, Catwalk is currently be used as the main evaluation framework for the OLMo project at AI2. As such, it has many features that allow it to be a replacement for other evaluation frameworks, such as Eleuther Harness \cite{eval-harness}. For more details about Catwalk for LLM evaluation, see \url{https://github.com/allenai/ai2-olmo-eval}.}

\section{Related Work}

Holistic Evaluation of Language Models (HELM; \citealp{LiangHELM2022}) also aims to increase the density of coverage between models and evaluations and take a complementary path to ours. HELM describes itself as a top-down approach; they design a taxonomy of scenarios and metrics and use them to curate a specific subset of combinations. Catwalk is bottom-up, and aims to lower the engineering barriers for model and task creators to integrate their work into a shared benchmark. For example, HELM chooses to only adapt models via a standardized 5-shot prompting method, while Catwalk explicitly incorporates different varieties of prompting and even fine-tuning.

Our work builds on the EleutherAI LM Evaluation Harness~\cite{eval-harness}. It similarly addresses the task of building a unified evaluation framework and focuses on decoder-only models and zero-/few-shot evaluations. By contrast, \catwalk also includes encoder-only and encoder/decoder models and supports fine-tuning.

\section{Conclusion}
We presented \catwalk, a system to evaluate many models on many datasets.
\catwalk's APIs eliminate redundant implementation efforts and unify incompatible approaches to language model evaluation.
The approach enables experimentation across datasets and language models at an unprecedented scale.
While a large number of models and datasets is already integrated, the Allen Institute for Artificial Intelligence continues to provide hands-on engineering help to grow this list, and invites further contributions as an open-source effort.

\section*{Limitations}
Language model evaluation can only be as good as the datasets that are used.
\catwalk takes no position on this topic, and instead makes as many datasets as possible available for study.
As shown in Table~\ref{tab:rank_correl}, the choice of dataset can have a big influence on which model appears ``best''.
Similarly, prompted models are often sensitive to the prompt, and fine-tuned models are sensitive to their training regime.
We hope that \catwalk will enable the kind of research that allows us to take strong, supported positions on these topics in the future.

\section*{Ethics Statement}

We believe the impact of this framework to be largely beneficial, in that it encourages the community to adopt reproducible research practices. A unified interface for evaluation allows for easy comparison across various papers, without the need for reimplementation and rerunning experiments that may take up significant time and compute resources. 

We also note that the datasets present in our framework may contain biases and artifacts of their own, and a model performing well within our framework on these datasets should not be considered as having “solved” the underlying task.

%\section*{Acknowledgements}

\bibliography{anthology,custom}

\clearpage
\appendix

\section{Appendix}
\label{sec:appendix}

\begin{sidewaystable*}[t]
\tiny
\begin{tabular}{r|r|r|r|r|r|r|r|r|r|r|r|r|r|r|r|r|r}
  & ARC Challenge & ARC Easy & CoLA & COPA & HEAD-QA EN & HellaSwag & LogiQA & MNLI & MRPC & OpenBookQA & PIQA & QNLI & QQP & RTE & SciQ & SST & WINOGRANDE\\
ARC Challenge & \cellcolor[rgb]{0.0,0.8,0.0}$1.000$ & \cellcolor[rgb]{0.13137599999999994,0.8262752,0.13137599999999994}$0.932$ & \cellcolor[rgb]{0.632764,0.9265528000000001,0.632764}$0.606$ & \cellcolor[rgb]{0.3422789999999999,0.8684558,0.3422789999999999}$0.811$ & \cellcolor[rgb]{0.3599999999999999,0.872,0.3599999999999999}$0.800$ & \cellcolor[rgb]{0.3177240000000001,0.8635448000000001,0.3177240000000001}$0.826$ & \cellcolor[rgb]{0.6554310000000001,0.9310862,0.6554310000000001}$0.587$ & \cellcolor[rgb]{0.487344,0.8974688000000001,0.487344}$0.716$ & \cellcolor[rgb]{0.587836,0.9175672,0.587836}$0.642$ & \cellcolor[rgb]{0.11827900000000013,0.8236558,0.11827900000000013}$0.939$ & \cellcolor[rgb]{0.623004,0.9246008,0.623004}$0.614$ & \cellcolor[rgb]{0.274096,0.8548192,0.274096}$0.852$ & \cellcolor[rgb]{0.515584,0.9031168,0.515584}$0.696$ & \cellcolor[rgb]{0.36638399999999993,0.8732768,0.36638399999999993}$0.796$ & \cellcolor[rgb]{0.705151,0.9410302,0.705151}$0.543$ & \cellcolor[rgb]{0.667071,0.9334142,0.667071}$0.577$ & \cellcolor[rgb]{0.441991,0.8883982,0.441991}$0.747$\\
\end{tabular}
\caption{Spearman correlations between models when ranked in order of performance on 17 common datasets.}
\label{tab:rank_correl}
\end{sidewaystable*}

\begin{sidewaystable*}[t]
\tiny
\begin{tabular}{l r r r r r r r r r r r r r r r r r r}
\toprule
& \bf Models & \bf ARC Challenge & \bf ARC Easy & \bf CoLA & \bf COPA & \bf HEAD-QA & \bf HellaSwag & \bf LogiQA & \bf MNLI & \bf MRPC & \bf OpenBookQA & \bf PIQA & \bf QNLI & \bf QQP & \bf RTE & \bf SciQ & \bf SST & \bf WINOGRANDE\\
% Ranked Classification Zero-shot\\
\midrule
\midrule
\parbox[t]{2mm}{\multirow{20}{*}{\rotatebox[origin=c]{90}{\bf Zero-shot Ranked Classification}}}
& BLOOM 560M & \cellcolor[rgb]{1,0.988120246287577,0.988120246287577}$-4.1$ & \cellcolor[rgb]{0.8249909453096704,1,0.8249909453096704}$60.4$ & \cellcolor[rgb]{0.9713147410358566,1,0.9713147410358566}$9.9$ & \cellcolor[rgb]{0.9304599782687433,1,0.9304599782687433}$24.0$ & \cellcolor[rgb]{0.8638174574429555,1,0.8638174574429555}$47.0$ & \cellcolor[rgb]{0.8762767113364723,1,0.8762767113364723}$42.7$ & \cellcolor[rgb]{0.9817457442955451,1,0.9817457442955451}$6.3$ & \cellcolor[rgb]{0.9814559942049982,1,0.9814559942049982}$6.4$ & \cellcolor[rgb]{0.8933719666787396,1,0.8933719666787396}$36.8$ & \cellcolor[rgb]{0.9327779789931184,1,0.9327779789931184}$23.2$ & \cellcolor[rgb]{0.9113364722926476,1,0.9113364722926476}$30.6$ & \cellcolor[rgb]{1,0.9985512495472655,0.9985512495472655}$-0.5$ & \cellcolor[rgb]{1,0.928141977544368,0.928141977544368}$-24.8$ & \cellcolor[rgb]{0.9843534951104672,1,0.9843534951104672}$5.4$ & \cellcolor[rgb]{0.4494748279608838,1,0.4494748279608838}$190.0$ & \cellcolor[rgb]{0.9110467222021007,1,0.9110467222021007}$30.7$ & \cellcolor[rgb]{0.9930459978268743,1,0.9930459978268743}$2.4$\\
\bottomrule
\end{tabular}
\caption{Improvement relative in percentage compared to the random baseline on 17 common tasks}
\label{tab:big_matrix}
\end{sidewaystable*}

% This is a section in the appendix.

\end{document}